\documentclass[conference]{IEEEtran}
\renewcommand\IEEEkeywordsname{Keywords}

\usepackage{comment}
\usepackage{graphicx,subfigure,url,balance,comment} 
\usepackage{algorithmic} 
\usepackage{multirow} 
\usepackage{amsmath} 
\usepackage{amsfonts} 
\usepackage{epstopdf} 
\usepackage{caption2}
\usepackage{upquote}
\usepackage{tabularx}
\usepackage{etoolbox}
\usepackage{color}
\usepackage[ruled]{algorithm2e}
\makeatletter
\patchcmd{\maketitle}{\@copyrightspace}{}{}{}
\makeatother

\usepackage{float}

\SetKwRepeat{doWhile}{do}{while}

\begin{document}
%

\title{A Deterministic Self-Organizing Map Approach and its Application on Satellite Data based Cloud Type Classification}

\author{%
	Wenbin Zhang$^1$, Jianwu Wang$^1$, Daeho Jin$^{2,3}$, Lazaros Oreopoulos$^3$, Zhibo Zhang$^4$ \\

  $^1$Department of Information Systems, University of Maryland, Baltimore County, Baltimore, MD, USA \\

  $^2$University Space Research Association, Columbia, MD, USA\\
  
  $^3$NASA Goddard Space Flight Center, Greenbelt, MD, USA\\
  
  $^4$Department of Physics, University of Maryland, Baltimore County, Baltimore, MD, USA\\
  
  \{wenbinzhang, jianwu, zhibo.zhang\}@umbc.edu, \{daeho.jin, lazaros.oraiopoulos-1\}@nasa.gov\\

}

%
%
%

%
%

\maketitle




\begin{abstract}
\label{sect:abstract}
A self-organizing map (SOM) is a type of competitive artificial neural network, which projects the high-dimensional input space  of the training samples into a low-dimensional space with the topology relations preserved. This makes SOMs supportive of organizing and visualizing complex data sets and have been pervasively used among numerous disciplines with different applications. Notwithstanding its wide applications, the self-organizing map is perplexed by its inherent randomness, which produces dissimilar SOM patterns even when being trained on identical training samples with the same parameters every time, and thus causes usability concerns for other domain practitioners and precludes more potential users from exploring SOM based applications in a broader spectrum. Motivated by this practical concern, we propose a deterministic approach as a supplement to the standard self-organizing map. In accordance with the theoretical design, the experimental results with satellite cloud data demonstrate the effective and efficient organization as well as simplification capabilities of the proposed approach. 

\end{abstract}

\vspace{2mm}

\section{Introduction}
\label{sect:intro}

The self-organizing map provides an automatic data analysis technique which helps to produce a low-dimensional representation, called map, of the high-dimensional input space without any external supervision \cite{polzlbauer2004survey}. Unlike in most biologically inspired neural network models, SOM performs competitive learning as opposed to error-correction learning by having units compete for the current object, and in the sense that it is topological preserving by having a neighborhood function to adjust the weights of neighbors of the winning unit concurrently. Various versions of the SOM models have been investigated over the years with practical applications across numerous disciplines, ranging from meteorology and oceanography to finance analysis, bioinformatics and image retrieval \cite{kohonen2013essentials, liu2011review}. In the pioneering work of the SOM-based meteorology application, Malmgren et al. showed the potential of such neural network system in identifying climate zones by organizing climate data, seasonal averages of precipitation and temperatures over the course of 30 years \cite{malmgren1999climate}. In \cite{iskandar2008impact}, the SOM was successfully applied to detect the dipole sea surface temperature anomaly pattern for Indian Ocean. Oja et al. used the SOM to study the mutual relationships of the HERVs and their similarities to other associated DNA elements \cite{oja2003clustering}. A tree structured SOM was designed to reduce the time complexity of search for the purpose of effective image retrieval and user's relevance feedback was interactively incorporated \cite{laaksonen1999picsom}. Including the just mentioned applications, the SOM-based approaches have gained their popularities as a powerful data analysis technique and have achieved varying degrees of success \cite{yin2008self}. 

One of the shortcomings of these methods, however, is the inherent randomness of SOM and the indeterministic arising therefrom \cite{liu2011review}, which perplexes SOM, results in dissimilar SOM patterns when being trained on the same training set with identical parameters and remains it as a black box especially to users without related background. As a result, dampening more potential users' interests in pursuing further SOM applications. Efforts have been made in offering suggestions and guidelines on how to handle the randomness of SOM, but it is desirable that the inherent randomness could be eliminated or at least minimized \cite{liu2006performance}. With this in mind, this paper proposes a variant deterministic self-organizing map to eliminate the randomness of standard self-organizing map approach. The maximum iterations parameter that was once necessary also becomes self-tuned as a byproduct of the random elimination process. These random eliminators are designed based on self-organizing map and are illustrated with an satellite cloud classification application, but they are generalizable knowledge in applications employing other learning algorithms. 

In summary, the contributions of this paper are:

\begin{itemize}
	\item A deterministic self-organizing map is proposed to eliminate the randomness of the standard self-organizing map. This deterministic network is invariant when the same training set is used.
	\item The proposed random eliminators are generalizable knowledge and are applicable for other learning algorithms. The tuning required maximum iterations parameter also becomes self-tuned during the randomness eliminating process. 
	\item The utilization of proposed network on real world satellite observation of clouds walks through the use of our method in practical application, which addresses the practical concerns of geoscientists and demonstrates the effectiveness and efficiency of our method.  
\end{itemize}

In the following sections, related studies will be firstly reviewed in Section~\ref{sect:related}. We  propose our method in Section~\ref{sec: determinizedApproach}. Section~\ref{sect:application} presents the application and then the performance will be analyzed and reasons for the results will be discussed in Section~\ref{sect:experiment}. Finally, Section~\ref{sect:conclusion} concludes the paper.

\vspace{-0mm}
\section{Related Work}
\label{sect:related}

\subsection{Deterministic Clustering}
\label{sec:related_deterministic}

Clustering algorithms are to partition objects into groups based on their similarity. Many clustering algorithms face indeterministic issue. For instance, as one of the mostly used clustering algorithm, the standard K-means algorithm~\cite{macqueen1967some, zhang2016using} randomly choose its initial centroids. It causes the algorithm to be very sensitive to its initial seed clusters and often produce very different results when running the K-means algorithm with the same parameter configuration. There have been several studies on how to achieve deterministic clustering~\cite{su2004deterministic, goder2008consensus}.~\cite{goder2008consensus} studies how to reconcile clustering results from different runs of the same algorithm and derive a consensus among them.~\cite{su2004deterministic} proposes to use principal component analysis (PCA) based divisive hierarchical approach for deterministic K-means initialization. This is also the mainstream in achieving deterministic clustering using SOM~\cite{akinduko2016som}. The high computational cost of determining the data-dependent PCA transform, however, hinders its applications in high-dimensional situations such as the classification of satellite cloud regimes.

\subsection{Satellite Data based Cloud Type Classification}
\label{sec:related_cloud}

The study of clouds, including their frequency of occurrence, location and characteristics, plays a key role in the understanding of climate change. Thick clouds in the lower atmosphere primarily reflect the incoming solar radiation and consequently cool the surface of the Earth. On the other hand, thin clouds in upper atmosphere easily transmit the incoming solar radiation and also trap some of the outgoing infrared radiation emitted by the EarthÕs surface and radiate it back downward, consequently warming the atmosphere and surface of the Earth. 

There have been many studies on cloud type classification. ~\cite{li2003high, li2007comparison} used maximum likelihood (ML) classification method to classify cloud types. In recent years, K-means, as one of the main approaches, has been widely used for cloud type clustering while others started to employ SOM for the cloud study. Our previous work~\cite{oreopoulos2014examination, oreopoulos2016radiative, Jin2017} used K-means approach to identify cloud regimes. As the pioneering work, McDonald et al. ~\cite{mcdonald2016automated} studied how to use SOM to identify cloud regimes and reported more objective organization compared to k-means. Because neither approach is deterministic, we still face usability challenges. This work is motivated by this critical practical concern. Collaborating with geoscientists, we aim to identify and interpret cloud regimes deterministically. 
\vspace{-0mm}
\section{The deterministic approach}
\label{sec: determinizedApproach}
\subsection{Standard Self Organizing Map}

An SOM \cite{kohonen1990self} is made up of a set of nodes. Each node holds a representative feature vector called the prototype. The standard SOM starts from randomly initializing the prototype feature vector of each node in the map. From there a sample vector is randomly selected and fed to the network, its Euclidean distances to all prototype vectors are then computed in order to find the neuron that most closely matches with the current sample vector. The prototype vector that best represents that sample becomes the winning unit and is called the Best Match Unit (BMU). Next, the neurons belong to the neighborhood set of BMU are also activated and the prototype vectors of all activated neurons are adjusted towards the input vector at the same time. From this step, the magnitude of the adjustment decreases with time and with distance from the BMU in an attempt to preserve topology relationships that exist within the input data. This whole process iterates until the predefined stopping condition is met. The common theme through the following sections is to eliminate the randomness of standard self-organizing map for the stable and efficient purpose, thus simplify the use of SOM for cloud regimes identification automata. A summary of notations used in this paper is given in Table~\ref{table:notation}.
 
 \begin{table}[!htb]
 	\centering
 	\setlength{\tabcolsep}{11pt}
 	\renewcommand\arraystretch{1.3}
 	\begin{tabular}{|c|c|c|}
 		\hline
 		Notation & Description \\
 		\cline{1-2} 
 		$\vec{\mkern1mu v}(t)$ & The prototype vector of each node in the map  at time \emph{t}\\
 		\cline{1-2}
 		$\vec{\mkern1mu f}$ & The current sample's feature vector  \\
 		\cline{1-2}
 		$BMU$ & \scriptsize The node that best matches with current sample's feature vector \\
 		\cline{1-2}
 		$L(0)$ & Initial learning rate  \\
 		\cline{1-2}
 		$L(t)$ & Learning rate at time \emph{t}  \\
 		\cline{1-2}
 		$R(0)$  & Initial neighborhood radius  \\
 		\cline{1-2}
 		$R(t)$  & Neighborhood radius at time \emph{t}  \\
 		\hline
 	\end{tabular}
 	\caption{Notation used for method description.}
 	\label{table:notation}
 \end{table}

\subsection{Update Procedure}
\label{sec:updateProcedure}
The implementation of self-organizing map algorithm demands the instantiation of its update procedure and there are two key components involved in the procedure: neighborhood radius and learning rate. The randomness roots in update procedure arises from the tunable parameters of distinct update functions as different parameters result in dissimilar SOM patterns. It is desirable that the tuning process could be eliminated or at least minimized \cite{liu2006performance}. Our approach therefore employs the update functions in which maximum iterations is the exclusive tunable parameter, and this parameter is self-tuned in the devised staggered sample selection method to be discussed in Section \ref{sec: sampleSelection}.

The first component neighborhood radius comes in a variety of flavors \cite{yin2008self}. Any node that is inside the neighborhood radius of a node that is to be updated also gets updated to a degree. Our approach uses the circle neighborhood that has an initial radius equal to half of the size of the smallest dimension of the SOM. Formally put:

\begin{equation}
\label{initialRadius}
R(0) = \frac{min(rows, colums)}{2}
\end{equation}

The radius of the SOM needs to decay with time so that the map will stabilize into its final organization. Here is the decay function used:
\begin{equation}
\label{radiusUpdate}
R(t)= R(0)\cdot b^{-{\frac{t}{\lambda}}}
\end{equation}

\noindent where $t = 0, 1, 2, \cdots, max(t)$, $\lambda=$ time constant $= \frac{max(t)}{\log_b R(0)}$ and $b=$ logarithmic base.

This decay function exponentially decays the original radius to 1 when \emph{t} reaches its maximum value, which is the maximum number of iterations which is specified in Section~\ref{sec: stoppingCondition}. If the SOM is continued to be trained at a radius of 1 the nodes immediately to the top, right, bottom, and left of the BMU would still be affected but this is not the case because the equation only becomes 1 when the maximum iteration is reached, meaning training is finished. This neighborhood function proves to work well as the amount that one node changes those around it decreases with time as it should \cite{flexer2001use}.

The learning rate is the amount of impact a sample that matches best to a node should have on that node and its neighbors. Like the neighborhood the learning rate similarly decays with time. Here is its function:

\begin{equation}
\label{learningRate}
L(t)= L(0)\cdot b^{-{\frac{t}{\lambda}}}
\end{equation}

\noindent where $t = 0, 1, 2, \cdots, max(t)$, $\lambda=$ time constant $= max(t)$ and $b=$ logarithmic base.

This is the same exponential decay function as the neighborhood radius except the time constant has changed to the given maximum number of iterations. The conducted experiments of this work show that speed and accuracy both increase as the initial learning rate decreases to a point. An initial rate of 0.1 works well. This is due to the fact that a high rate causes more oscillation in the prototype vectors of the SOM nodes because as two or more samples may jockey for main position in the same BMU. A high learning rate also causes the node to swing more toward the most recently matching sample than it should, meaning it ``forgets" the impact of the other nodes that matched to it before too quickly. 

There is another factor that adjusts the learning rate of a neighborhood node ($i, j$) of a BMU ($k, l$) by taking the actual distance that the neighboring node is from the BMU into account. Here is the mathematical definition of influence used in our method, note that it also decays with time:

\begin{equation}
\label{influence}
	I(t)= b^{-{\frac{d_{(i,j)(k,l)}}{2R(t)}}}
\end{equation}
\noindent where $t= 0, 1, 2, \cdots, max(t)$, $I(t)=$ influence on node (\emph{i, j}) by (\emph{k, l}) at time \emph{t}, $d_{(i, j)(k, l)}=$ distance from node (\emph{i, j}) to (\emph{k, l}) and $b=$ logarithmic base.

As is evident the influence exponentially decreases the further the neighboring node is from the BMU using the ratio between the distance between the two nodes and the current neighborhood diameter. 

All of the pieces are combined into a single vector update equation, formally put:

\begin{equation}
\label{equation: updateEquation}
\vec{\mkern1mu v}(t+1)= \vec{\mkern1mu v}(t)+ I(t)L(t)(\vec{\mkern1mu f}-\vec{\mkern1mu v}(t))
\end{equation}

\subsection{Stopping Condition}
\label{sec: stoppingCondition}
Considering that the speed at which the SOM can be trained hinges on the stopping condition, we consider the two types of convergence mechanisms in our approach. The first type of stopping condition simply uses a user specified maximum iterations parameter as the upper bound on the training iterations. An iteration is the whole training set being shown to the SOM. The maximum number of training times is based on the given parameter and then the training stops. The drawback to this approach is that it does not take the map activity into account therefore training may proceed through many iterations while the SOM is actually producing no improvement \cite{yin2008self}. In this case if the system had some measurement of improvement after each iteration it could stop earlier when it sees that there was no improvement can be made.

The second type stopping condition, namely No Moves, defines ``no improvement" in SOM's status as no training samples changing their best match unit in a complete iteration of the training set \cite{yin2008self}. Using this condition the training process is stopped as soon as it sees no improvement. While this stopping condition could be used alone in some SOMs our approach uses it in conjunction with the maximum iterations condition described above. It is because, as described in Section~\ref{sec:updateProcedure} on the proposed method's updating procedure, the learning rate and neighborhood radius update equations require maximum number iterations to be known. In any event, using this condition alone could also result in infinite training if update factors are not set up to decay correctly. For this reason the maximum iterations condition can be think of hidden in the background as a safety net that will rarely being need, but it is nice to have there just in case. It has to be said that while this stopping condition, as a proxy for the true goal, might not be fully optimized just like other heuristic strategies, a local optimum returned by simple greedy search may be better than the global optimum \cite{domingos2012few}. The study of the SOM's behavior from this work also has shown that the point where the samples stop changing their BMU is the point where map configuration has often peaked. On the occasion beyond ``often", it is the chance that it has not and as such user runs the risk when using this method that user may miss out on a slightly better map organization, it is really matter whether user wants to wait through the maximum number iterations or not for a little improvement, if any. There is another small caveat that developed through this study's observing of the SOM's activities. If this stopping condition is used and the training usually halts before the specified maximum iterations is reached. Although samples are distributed correctly the actual prototype vector of the node may not reflect the samples as well as it should. Recall that a SOM works on the ``best match principle'' and so although the prototype vector may not represent the samples in its node well it can still be the best match for those samples when compared with the rest of the nodes in the map. A figured out way to cope with this is to turn the initial learning rate up a little. Although this flies in the face of what is found to work best in the general case in the learning rate section above, it tends to produce slightly better map configuration in some cases if the map does converge too fast under this condition. The reason for this is that if the map converges too fast the number of iterations is small and as such given that the training samples are only shown to the SOM a short few times they need to impact their best match unit's prototype vector fast so that any sample shown to the SOM will find the prototype vectors containing the other samples it usually should.

\subsection{Initialization Method}
Initializing a node refers to setting the values in the prototype vector of that node before training begins. Random initialization is the common technique used to initialize the nodes. It simply means to run through every value in the nodes' vectors that need to be initialized and set them to a random value. It was found that this technique is good for producing an even distribution but because of the randomness a random distribution will be produced every time \cite{akinduko2016som}. Even though the cloud data samples will organize nicely once there has been enough iteration chances are they will be in different cloud type nodes when the training completes. The layout of these different nodes may be better or worse than the other times, there is no guarantee due to the randomness. To eliminate the initialization randomness of standard SOM method, we set up the nodes with a smooth transition from the top left to the bottom right corner. The gradient initialization computes the initial values of the prototype vector of any node ($a, b$) in the following way:

\begin{equation}
v_i = \frac{d_{(1,1)(a,b)}}{d_{(1,1)(m,n)}}
\end{equation}

\noindent where $d_{(1, 1)(m, n)}$ is the maximum distance possible of the map, that is the distance from the top left node to the bottom right node.

\subsection{Sample Selection}
\label{sec: sampleSelection}
It is important with most, if not all, neural networks that during the training process the samples are not fed to the network sequentially in the same order every time. Doing this may cause a bias for either the beginning or the ending input samples, depending on the type of network \cite{zadrozny2004learning}. Random selection is the commonly used technique. As its name suggests, available samples are randomly selected from the training set during sample selection step of standard SOM. The consequences of this randomness are that while a good distribution should be achieved it may not always be, and not only that, it is hard to tell one training run from another since nodes' sets of samples, while normally similar, will be shifted around. The organization will thus be randomly produced each time with this method \cite{yin2008self}. To come up with consistent results a staggered selection method that tries to maintain the good characteristics of randomness and eliminate the actual randomness is devised. The staggered approach gives all of the training samples equal opportunity to start an iteration at some point during the training. As well as giving the samples this equal opportunity it ensures that the samples are not shown to the learning algorithm in the same order during each training iteration. The idea is detailed in Algorithm~\ref{alg:staggered}.

\begin{algorithm}
	\label{alg:staggered}
	\caption{Staggered sample selection algorithm}
	\LinesNumbered 
	\KwIn{Training samples' index.}
	\KwOut{Training order list;\\ \ \ \ \ \ \ \ \ \ \ \ Maximum iterations of SOM.}
	
	\While{$frontIndex \leq backIndex$}{
		\eIf{$reverse$}{$startIndex= backIndex$\;}{$startIndex= frontIndex$\;}
		$curretIndex= startIndex$\;
		
		\doWhile{		
			$currentIndex \ != startIndex$
		}{
			Train the network on the sample at $currentIndex$\;
			
			\eIf{$reverse$}{
				$currentIndex = (currentIndex -1 + S_{sample})\ \% \ S_{sample}$\;
			}{
				$currentIndex = (currentIndex +1 + S_{sample})\ \% \ S_{sample}$\;
			}
	
		}
	
	\eIf{reverse}{$backIndex--$\;}{$frontIndex++$\;}
	$reverse \ =! \ reverse$\;
	$maxIteration++$\;
	}
\end{algorithm}

In Algorithm~\ref{alg:staggered}, a front index ($frontIndex$) and a back index ($backIndex$) that point to the first element and last element in the list of training samples are initialized respectively. The current direction ($reverse$) is initialized as $false$, which means the current direction is set to forward. Another two indexes, start index $(startIndex)$ and current index ($currentIndex$) refer to the initial element and the current element respectively. The staggered sample selection first (lines 2-6) determines the sample that bootstraps each iteration during the training. If the current direction is forward the start index ($startIndex$) is set equal to the front index, otherwise, i.e., when $reverse$ is $true$, set the start index equal to the back index. Lines 8-15 complete one whole training iteration. The network is first trained on the sample that the current index points to. Next, if the current direction is forward increment the current index until the current index is equal to the front index. In the meanwhile, any change in the current index that goes outside the range of the training list wraps around to the other side. The same scheme applies when the current direction is reverse but the current index works in decrement fashion until the current index is equal to the back index. Lines 16-20 update the front index or back index after each iteration depending on the current direction. If the current direction is forward increment the front index, otherwise decrement the back index. Finally the current direction is reversed per iteration as well, that is to say if the current direction is forward then set it to reverse, otherwise set the current direction to forward. The algorithm ceases when the front index is greater than the back index.

This staggered method of selection produces equivalent results to its random counterpart every time. This equivalence is guaranteed because the results are not random and therefore have the same performance every time. In addition, by changing the start sample and reversing the order of input after each iteration, this method has the effect of evening out the influence of a single training sample because right after a sample is used the first becomes the last, and then second, and then second last, and so on. Another convenience of this method is that the maximum number of iterations is chosen by the size of the training list alone making the maximum iterations parameter that was once necessary obsolete. In summary, this staggered method of selection is relatively fast, consistent, and tunes the maximum iterations parameter, which is required in the update equations and stopping condition.

\subsection{Towards Self-tuned}
\label{sec: towardsSelf-tuned}

Despite its gained popularity as a powerful data analysis technique in a variety of communities, the self-organizing map remains as a black box especially to users without related background due to its parameter choices. Although efforts have been made in providing guidelines on how to tune the SOM, the distinct choices of tunable parameters may result in dissimilar SOM patterns. It is thus anticipated that the parameter choices could towards self-tuned in order to further streamline the usage of self-organizing map based approaches \cite{liu2006performance}. In our proposed approach, an SOM can be produced by supplying just two parameters, the average samples per node desired (from which the SOM dimension can be derived) and the initial learning rate of the SOM. The initial neighborhood radius, radius decay function, learning rate function as well as the maximum iterations parameter are self-tuned during the randomness eliminating process, which minimize the tuning effort thus simplify the use of SOM for geoscience as well as other domain practitioners.

\vspace{-0mm}
\section{Application}
\label{sect:application}

\begin{figure}[ht]
    \centering
    \includegraphics[width=0.45\textwidth]{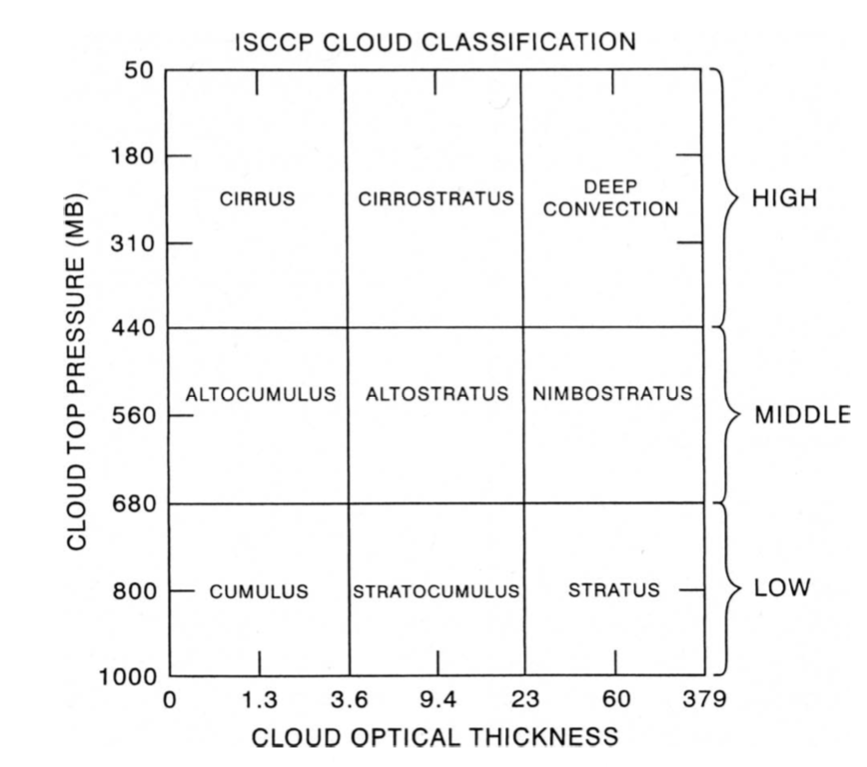}
    \caption{Assignment of traditional cloud types to 2-D joint histogram of COT and CTP (so-called ISCCP-like 2-D histogram)~\cite{rossow1999advances}.}
    \label{fig:my_label}
\end{figure}

The study of clouds, including their frequency of occurrence, location and characteristics, plays a key role in the understanding of climate variability and climate change. Clouds have complex impacts on the EarthÕs climate since they interact with both the incoming solar radiation and outgoing infrared radiation, with the interactions depending strongly on cloud altitude and thickness. In this sense, cloud optical thickness (COT) and cloud top pressure (CTP) are key variables for describing both the solar and infrared radiative effects of cloud. A data set of passive cloud retrievals, the International Satellite Cloud Climatology Project (ISCCP) employed these two variables to build a 2-D joint histogram~\cite{rossow1999advances}, shown in Figure~\ref{fig:my_label}, to distinguish among different cloud types with distinct radiative effects.



The 2-D joint histograms of satellite cloud retrievals have proven to be a useful dataset to perform and study cloud classification. Because both optical thickness and top pressure of cloud may vary significantly in the scale of $O$(100km), a rather expansive 2-D COT-CTP joint histogram is required to describe co-variation of COT and CTP. The ISCCP-like MODIS 2-D joint histogram consists of 42 elements (= 6 classes of COT $\times$ 7 classes of CTP), with each element representing the occurrence of a specific COT-CTP combination as cloud fraction (CF) ranging from 0 to 1. A big scientific challenge is to group the satellite images represented by the 2-D joint histograms into different clusters, one example of which is the ``Cloud Regime" (CR)~\cite{oreopoulos2016radiative, Jin2017}.

\begin{figure}
    \centering
    \includegraphics[width=0.45\textwidth]{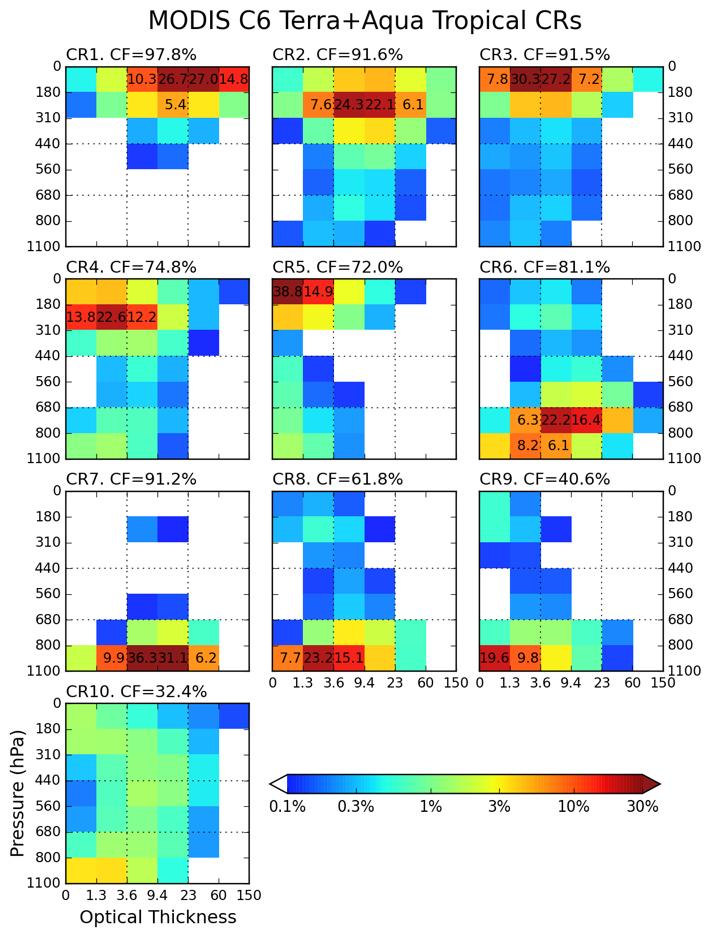}
    \caption{The cloud regime (CR) centroids of daily ISCCP joint histograms. 
    The cloud fraction (CF) of each regime, namely sum of 42 bin CF values, is also provided.}
    \label{fig:multHist}
\end{figure}

The ``Cloud Regime" is a concept of dominant mixtures of cloud types represented by the means of similar co-variations of 2-D joint histogram. Our previous works in~\cite{oreopoulos2014examination, oreopoulos2016radiative} obtained CR from K-means clustering analysis~\cite{macqueen1967some}. Figure~\ref{fig:multHist} shows one of optimal centroid of K-means clustering using the same data set at the same region (tropics; see Section~\ref{sect:Dataset} for details of data set). It shows that tropical cloud variability can be explained by 5 high cloud regimes, 4 low cloud regimes, and 1 semi-clear regime with quite low CF. The relative frequency of occurrence (RFO) map indicates that the high cloud regimes usually occur over the tropical warm pool area, intertropical convergence zone (ITCZ), and land area, while low and thick clouds crowds eastern side of oceans (not shown). The semi-clear regime is very popular all over the tropics. 


\vspace{-0mm}
\section{Experiments}
\label{sect:experiment}

This section analyzes how specific details of the proposed deterministic network affects its map configuration and running time when it is applied for satellite cloud classification.

\subsection{Dataset}
\label{sect:Dataset}


In this experiment, we used 2-D joint histogram of COT and CTP from the Moderate Resolution Imaging Spectroradiometer (MODIS) instrument aboard the Aqua satellite. The MODIS cloud data set (MYD08 D3~\cite{king2003cloud, platnick2003modis}) provides Level-3 cloud products at daily timescales with 1$^{\circ}$ $\times$ 1$^{\circ}$ horizontal resolution. We used the latest version of the MODIS atmospheric data sets, ``Collection 6"~\cite{platnick2017modis}.  
Specifically, we used Level-3 2-D joint histogram in the tropics (15$^{\circ}$S - 15$^{\circ}$N) for one year (2005). Thus the input dimension is 42 array elements, 360 $\times$ 30 spatial elements (grid cells), and 365 days. For the clustering analysis, missing data and completely cloud-free data (all 42 values are zero) are excluded. There are 3,445,612 records in total.

\subsection{Cloud Classification Result}

\begin{figure}
\centering  
\subfigure[Standard SOM execution result 1.]{\includegraphics[width=0.9\linewidth]{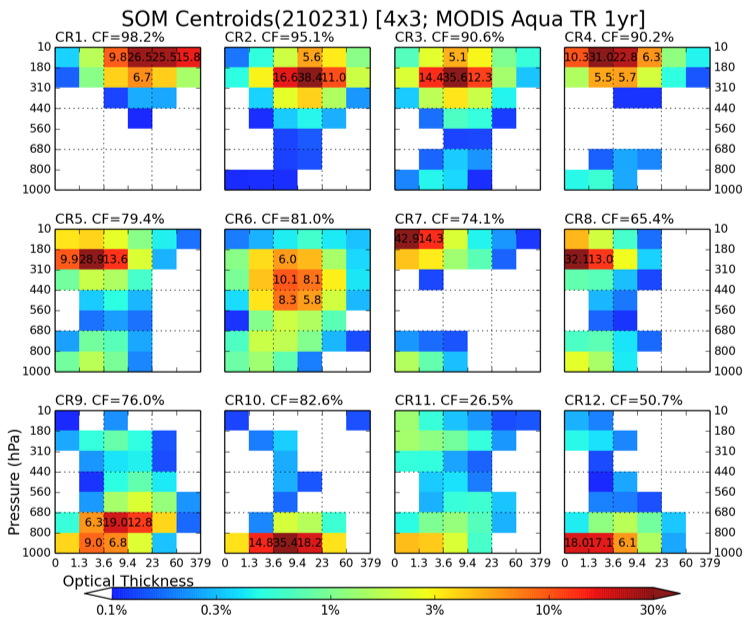}}
\subfigure[Standard SOM execution result 2.]{\includegraphics[width=0.9\linewidth]{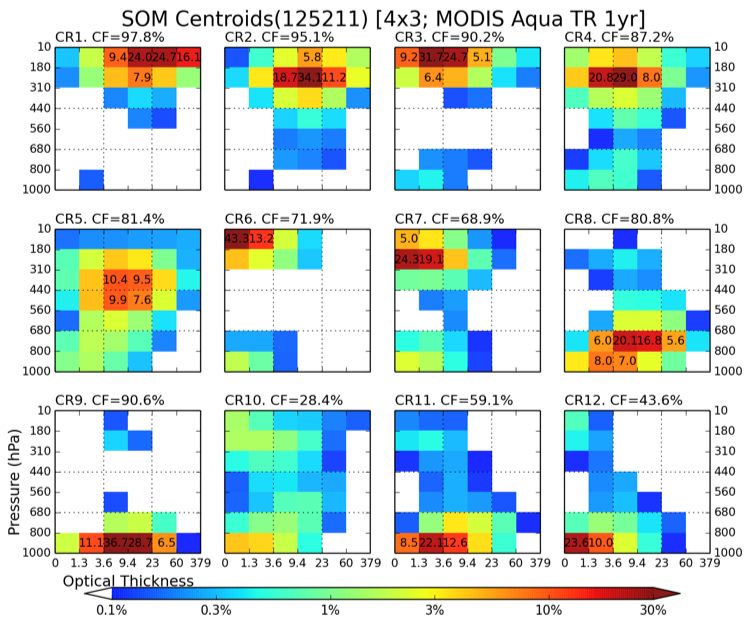}}
\subfigure[Standard SOM execution result 3.]{\includegraphics[width=0.9\linewidth]{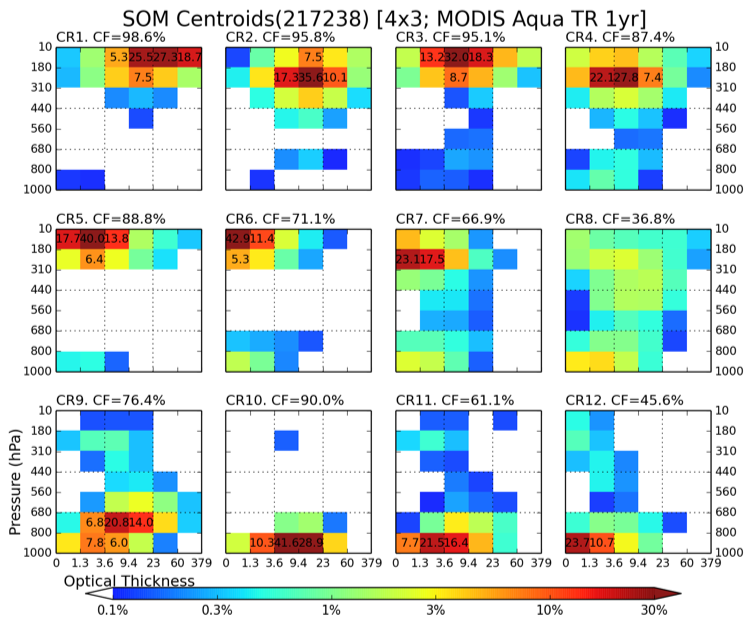}}
\vspace{-1mm}
\caption{The SOM cloud type vectors displayed as joint histograms by three standard SOM execution results using the same parameter configuration.}
\label{fig:standardSOM}
\end{figure}

\begin{figure}
\centering  
\subfigure[Standard SOM execution result 1.]{\includegraphics[width=0.9\linewidth]{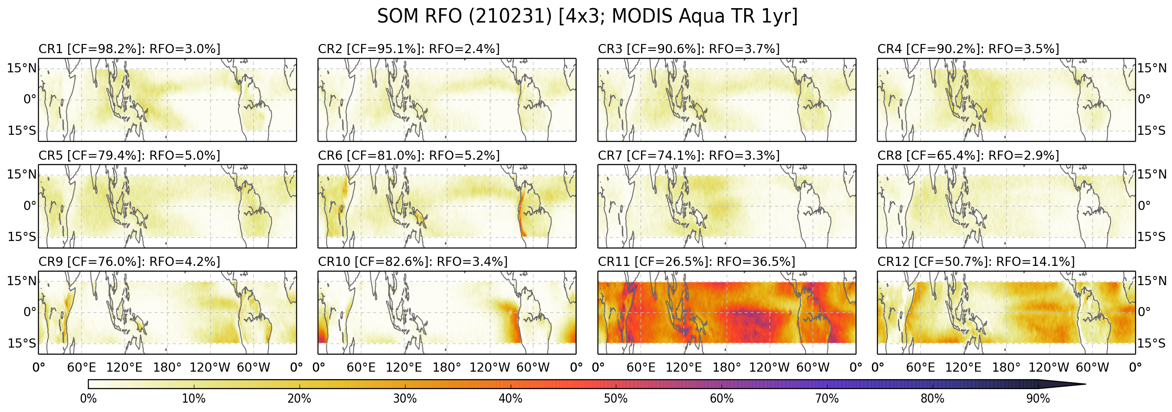}}
\subfigure[Standard SOM execution result 2.]{\includegraphics[width=0.9\linewidth]{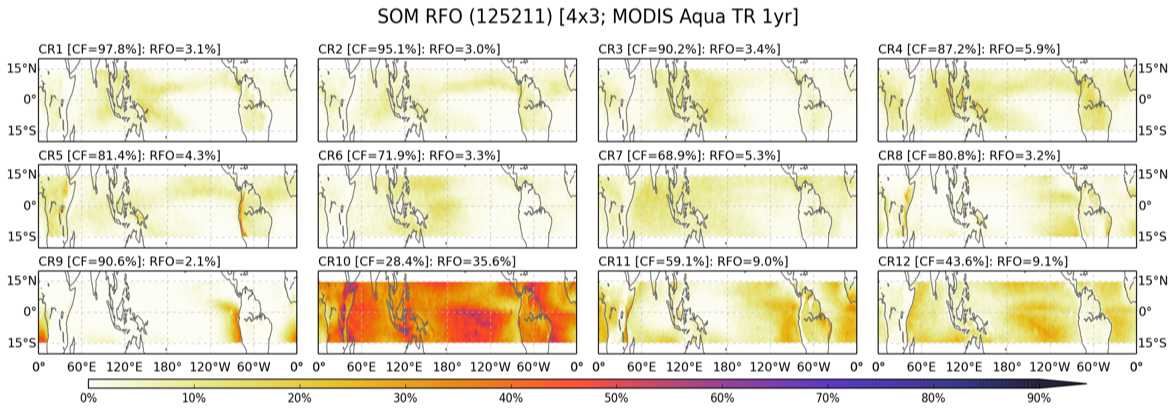}}
\subfigure[Standard SOM execution result 3.]{\includegraphics[width=0.9\linewidth]{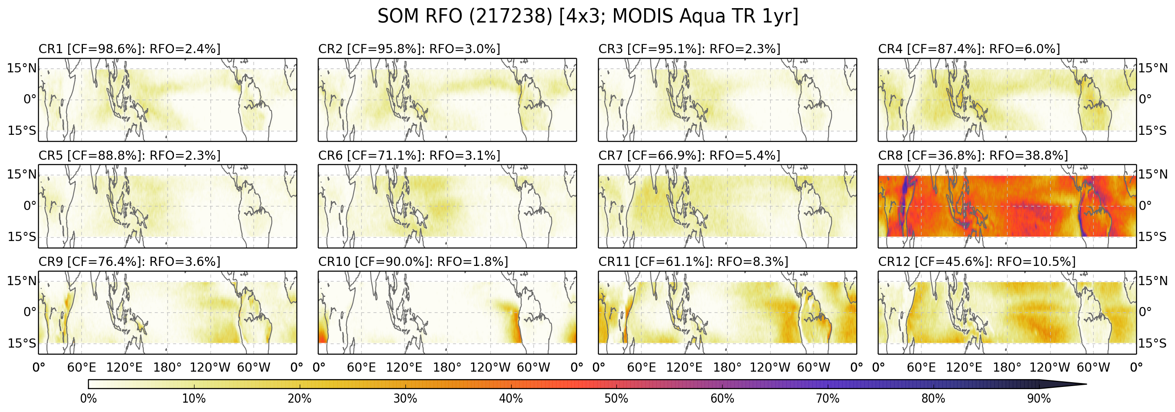}}
\vspace{-1mm}
\caption{The relative frequency of occurrence (RFO) corresponding to Figure~\ref{fig:standardSOM}.}

\label{fig:standardSOM-RFO}
\end{figure}

\begin{figure}
  \vspace{-5mm}
    \centering
    \includegraphics[width=0.45\textwidth]{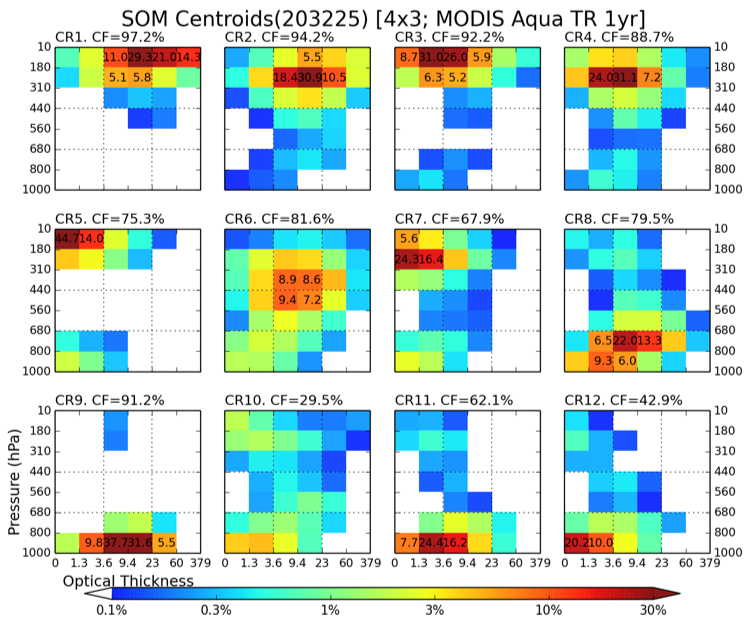}
    \vspace{-1mm}
    \caption{The SOM cloud type vectors displayed as joint histograms using proposed deterministic SOM.}
    \label{fig:joinHistogram}
\end{figure}

\begin{figure}
    \vspace{-6mm}
    \centering
    \includegraphics[width=0.45\textwidth]{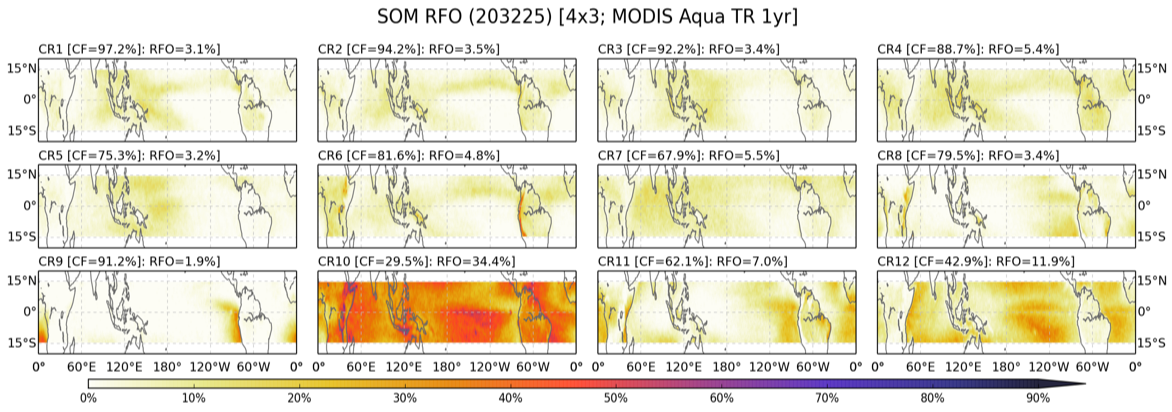}
    \vspace{-1mm}
    \caption{The relative frequency of occurrence (RFO) corresponding to Figure~\ref{fig:joinHistogram}.}
    \label{fig:RFO}
\end{figure}

We first investigate the map configuration of our proposed network w.r.t. determinateness and cloud regimes classification. A 4$\times$3 SOM was selected as suggested by~\cite{mcdonald2016automated} and initial learning rate was set as 0.1 for all map experimentation. Figure~\ref{fig:standardSOM} and Figure~\ref{fig:standardSOM-RFO} show the CR joint histograms and the associated relative frequency of occurrence (RFO) map associated with each node in the SOM from multiple executions of the same standard SOM with the same parameter configurations. It clearly shows that the standard SOM produces different CR histograms when trained with the same training set and identical parameters. On the contrary, using the proposed algorithm, the results, shown in Figure~\ref{fig:joinHistogram} and Figure~\ref{fig:RFO}, are invariant when the same training set and dimension are used. This confirms that our proposed method of random eliminators produces consistent and predictable results in terms of physical sense.




Compared to the K-means results shown in Figure~\ref{fig:multHist}, the SOM results produce reasonable CRs despite the different configuration of total number of CRs. For example, the CR histograms of deterministic SOM result (Figure~\ref{fig:joinHistogram}) contain all of CR histogram characteristics shown in Figure~\ref{fig:multHist}. In the case of standard SOM results (Figure~\ref{fig:standardSOM}), the CR output is slightly unsatisfactory because CR2 and CR3 here shares large similarity in both histogram and RFO map patterns, correlation coefficients of which are 0.60 and 0.74, respectively. This result shows that our deterministic SOM algorithm produces quality CRs.


\subsection{Execution Time}
The running time of the satellite cloud classification task is mainly dominant by the time for training and classifying. Specific to SOM, after the training is complete, each processing cloud data is labeled with associated cloud type. The training speed is therefore responsible for the main execution time difference. In this experiment, we verify the efficiency of two proposed random eliminators in determinizing SOM. These two eliminators include (1) the use of gradient initialization to compute the initial values of the prototype vector of cloud type node (denoted as GI); (2) staggered sample selection to feed cloud data for training (denoted as SSS). The performance difference of adding each eliminator is shown in Table~\ref{table:componentsComp}.

\begin{table}[!htb]
	\centering
	\setlength{\tabcolsep}{11pt}
	\renewcommand\arraystretch{1.3}
	\begin{tabular}{|c|c|c|c|}
		\hline
		Eliminators & SOM & SOM+GI & SOM+GI+SSS\\
		\cline{1-4}
		Time & 41 & 36 & 35 \\
		\hline
	\end{tabular}
	\caption{Running time (minutes) comparison of adding each eliminator for satellite cloud classification (GI: gradient initialization, SSS: staggered sample selection, SOM: standard SOM).}
	\label{table:componentsComp}
\end{table}

Table~\ref{table:componentsComp} shows our proposed network, other than the desired deterministic property, also produces its results in a more efficient manner as a bonus. Sufficed to say, this is because after initializing the network gradiently, a pattern is already present so the nodes organize around it more quickly than having to jostle randomly initialized ones into a pattern while they organize themselves. The further inclusion of staggered sample selection incurs no extra runtime costs but guarantees the map configuration is not random and maintains the good characteristics of random selection at the same time.

\vspace{-0mm}
\section{Conclusion and Future Work}
\label{sect:conclusion}

This work is motivated by the usability concern caused by inherent randomness of SOM. To address this practical concern, we propose a deterministic self-organizing map with effective satellite cloud type organization and execution capabilities. The improvements by including the concocted random eliminators are that not only both the speed and clustering of the training improved but by running with identical training samples the same resultant SOM will be produced every time. These random eliminators are generalizable knowledge and can be used as supplements to other iterative methods of learning neural networks. We successfully applied our deterministic SOM to a real-world scientific application to demonstrate its effectiveness and efficiency. It is anticipated that the proposed deterministic SOM could simplify the usage of self-organizing map based approaches and stimulate more potential user's interests in pursuing further SOM applications. In the future, we plan to extend the proposed network in conjunction with our previous work~\cite{zhang2017hybrid} for streaming scenarios.

\section{Acknowledgment}

This work is supported by the grant CyberTraining: DSE: Cross-Training of Researchers in Computing, Applied Mathematics and Atmospheric Sciences using Advanced Cyberinfrastructure Resources from the National Science Foundation (grant no.\ OAC--1730250).

\vspace{-0mm}



%
\bibliographystyle{abbrv}
\bibliography{ref}

\begin{thebibliography}{10}

\bibitem{akinduko2016som}
A.~A. Akinduko, E.~M. Mirkes, and A.~N. Gorban.
\newblock Som: Stochastic initialization versus principal components.
\newblock {\em Information Sciences}, 364:213--221, 2016.

\bibitem{domingos2012few}
P.~Domingos.
\newblock A few useful things to know about machine learning.
\newblock {\em Communications of the ACM}, 55(10):78--87, 2012.

\bibitem{flexer2001use}
A.~Flexer.
\newblock On the use of self-organizing maps for clustering and visualization.
\newblock {\em Intelligent Data Analysis}, 5(5):373--384, 2001.

\bibitem{goder2008consensus}
A.~Goder and V.~Filkov.
\newblock Consensus clustering algorithms: Comparison and refinement.
\newblock In {\em Proceedings of the Meeting on Algorithm Engineering \&
  Expermiments}, pages 109--117. Society for Industrial and Applied
  Mathematics, 2008.

\bibitem{iskandar2008impact}
I.~Iskandar, T.~Tozuka, Y.~Masumoto, and T.~Yamagata.
\newblock Impact of indian ocean dipole on intraseasonal zonal currents at 90 e
  on the equator as revealed by self-organizing map.
\newblock {\em Geophysical Research Letters}, 35(14), 2008.

\bibitem{Jin2017}
D.~Jin, L.~Oreopoulos, and D.~Lee.
\newblock Regime-based evaluation of cloudiness in cmip5 models.
\newblock {\em Climate Dynamics}, 48(1):89--112, Jan 2017.

\bibitem{king2003cloud}
M.~D. King, W.~P. Menzel, Y.~J. Kaufman, D.~Tanr{\'e}, B.-C. Gao, S.~Platnick,
  S.~A. Ackerman, L.~A. Remer, R.~Pincus, and P.~A. Hubanks.
\newblock Cloud and aerosol properties, precipitable water, and profiles of
  temperature and water vapor from modis.
\newblock {\em IEEE Transactions on Geoscience and Remote Sensing},
  41(2):442--458, 2003.

\bibitem{kohonen1990self}
T.~Kohonen.
\newblock The self-organizing map.
\newblock {\em Proceedings of the IEEE}, 78(9):1464--1480, 1990.

\bibitem{kohonen2013essentials}
T.~Kohonen.
\newblock Essentials of the self-organizing map.
\newblock {\em Neural networks}, 37:52--65, 2013.

\bibitem{laaksonen1999picsom}
J.~Laaksonen, M.~Koskela, and E.~Oja.
\newblock Picsom: Self-organizing maps for content-based image retrieval.
\newblock In {\em Neural Networks, 1999. IJCNN'99. International Joint
  Conference on}, volume~4, pages 2470--2473. IEEE, 1999.

\bibitem{li2003high}
J.~Li, W.~P. Menzel, Z.~Yang, R.~A. Frey, and S.~A. Ackerman.
\newblock High-spatial-resolution surface and cloud-type classification from
  modis multispectral band measurements.
\newblock {\em Journal of Applied Meteorology}, 42(2):204--226, 2003.

\bibitem{li2007comparison}
Z.~Li, J.~Li, W.~P. Menzel, T.~J. Schmit, and S.~A. Ackerman.
\newblock Comparison between current and future environmental satellite imagers
  on cloud classification using modis.
\newblock {\em Remote Sensing of Environment}, 108(3):311--326, 2007.

\bibitem{liu2011review}
Y.~Liu and R.~H. Weisberg.
\newblock A review of self-organizing map applications in meteorology and
  oceanography.
\newblock In {\em Self Organizing Maps-Applications and Novel Algorithm
  Design}. InTech, 2011.

\bibitem{liu2006performance}
Y.~Liu, R.~H. Weisberg, and C.~N. Mooers.
\newblock Performance evaluation of the self-organizing map for feature
  extraction.
\newblock {\em Journal of Geophysical Research: Oceans}, 111(C5), 2006.

\bibitem{macqueen1967some}
J.~MacQueen et~al.
\newblock Some methods for classification and analysis of multivariate
  observations.
\newblock In {\em Proceedings of the fifth Berkeley symposium on mathematical
  statistics and probability}, volume~1, pages 281--297. Oakland, CA, USA,
  1967.

\bibitem{malmgren1999climate}
B.~A. Malmgren and A.~Winter.
\newblock Climate zonation in puerto rico based on principal components
  analysis and an artificial neural network.
\newblock {\em Journal of climate}, 12(4):977--985, 1999.

\bibitem{mcdonald2016automated}
A.~J. McDonald, J.~J. Cassano, B.~Jolly, S.~Parsons, and A.~Schuddeboom.
\newblock An automated satellite cloud classification scheme using
  self-organizing maps: Alternative isccp weather states.
\newblock {\em Journal of Geophysical Research: Atmospheres}, 121(21), 2016.

\bibitem{oja2003clustering}
M.~Oja, P.~Somervuo, S.~Kaski, and T.~Kohonen.
\newblock Clustering of human endogenous retrovirus sequences with median
  self-organizing map.
\newblock In {\em Proc. WSOM}, volume~3, 2003.

\bibitem{oreopoulos2016radiative}
L.~Oreopoulos, N.~Cho, D.~Lee, and S.~Kato.
\newblock Radiative effects of global modis cloud regimes.
\newblock {\em Journal of Geophysical Research: Atmospheres},
  121(5):2299--2317, 2016.

\bibitem{oreopoulos2014examination}
L.~Oreopoulos, N.~Cho, D.~Lee, S.~Kato, and G.~J. Huffman.
\newblock An examination of the nature of global modis cloud regimes.
\newblock {\em Journal of Geophysical Research: Atmospheres},
  119(13):8362--8383, 2014.

\bibitem{platnick2003modis}
S.~Platnick, M.~D. King, S.~A. Ackerman, W.~P. Menzel, B.~A. Baum, J.~C.
  Ri{\'e}di, and R.~A. Frey.
\newblock The modis cloud products: Algorithms and examples from terra.
\newblock {\em IEEE Transactions on Geoscience and Remote Sensing},
  41(2):459--473, 2003.

\bibitem{platnick2017modis}
S.~Platnick, K.~G. Meyer, M.~D. King, G.~Wind, N.~Amarasinghe, B.~Marchant,
  G.~T. Arnold, Z.~Zhang, P.~A. Hubanks, R.~E. Holz, et~al.
\newblock The modis cloud optical and microphysical products: Collection 6
  updates and examples from terra and aqua.
\newblock {\em IEEE Transactions on Geoscience and Remote Sensing},
  55(1):502--525, 2017.

\bibitem{polzlbauer2004survey}
G.~P{\"o}lzlbauer.
\newblock {\em Survey and comparison of quality measures for self-organizing
  maps}.
\newblock na, 2004.

\bibitem{rossow1999advances}
W.~B. Rossow and R.~A. Schiffer.
\newblock Advances in understanding clouds from isccp.
\newblock {\em Bulletin of the American Meteorological Society},
  80(11):2261--2288, 1999.

\bibitem{su2004deterministic}
T.~Su and J.~Dy.
\newblock A deterministic method for initializing k-means clustering.
\newblock In {\em Tools with Artificial Intelligence, 2004. ICTAI 2004. 16th
  IEEE International Conference on}, pages 784--786. IEEE, 2004.

\bibitem{yin2008self}
H.~Yin.
\newblock The self-organizing maps: background, theories, extensions and
  applications.
\newblock In {\em Computational intelligence: A compendium}, pages 715--762.
  Springer, 2008.

\bibitem{zadrozny2004learning}
B.~Zadrozny.
\newblock Learning and evaluating classifiers under sample selection bias.
\newblock In {\em Proceedings of the twenty-first international conference on
  Machine learning}, page 114. ACM, 2004.

\bibitem{zhang2016using}
W.~Zhang, J.~Tang, and N.~Wang.
\newblock Using the machine learning approach to predict patient survival from
  high-dimensional survival data.
\newblock In {\em Bioinformatics and Biomedicine (BIBM), 2016 IEEE
  International Conference on}, pages 1234--1238. IEEE, 2016.

\bibitem{zhang2017hybrid}
W.~Zhang and J.~Wang.
\newblock A hybrid learning framework for imbalanced stream classification.
\newblock In {\em Big Data (BigData Congress), 2017 IEEE International Congress
  on}, pages 480--487. IEEE, 2017.

\end{thebibliography}
%
%
\end{document}